\documentclass[12pt]{article}%
\usepackage{amsmath}
\usepackage{graphicx,float,wrapfig}
\usepackage{amsfonts}
\usepackage{amssymb}
\usepackage{algorithm}
\usepackage{algorithmic}
\usepackage[T1]{fontenc}
\usepackage{graphicx}%
\providecommand{\U}[1]{\protect\rule{.1in}{.1in}}
\begin{document}

\title{An Adaptive Weighted Deep Forest Classifier}
\author{Lev V. Utkin$^{1}$, Andrei V. Konstantinov$^{1}$, Viacheslav S. Chukanov$^{1}$,
\and Mikhail V. Kots$^{1}$, Anna A. Meldo$^{1,2}$\\$^{1}${\small Peter the Great St.Petersburg Polytechnic University (SPbPU)}\\$^{2}${\small St.Petersburg Clinical Research Center for Special Types of
Medical Care (Oncology-oriented)}}
\date{}
\maketitle

\begin{abstract}
A modification of the confidence screening mechanism based on adaptive
weighing of every training instance at each cascade level of the Deep Forest
is proposed. The idea underlying the modification is very simple and stems
from the confidence screening mechanism idea proposed by Pang et al. to
simplify the Deep Forest classifier by means of updating the training set at
each level in accordance with the classification accuracy of every training
instance. However, if the confidence screening mechanism just removes
instances from training and testing processes, then the proposed modification
is more flexible and assigns weights by taking into account the classification
accuracy. The modification is similar to the AdaBoost to some extent.
Numerical experiments illustrate good performance of the proposed modification
in comparison with the original Deep Forest proposed by Zhou and Feng.


\end{abstract}

\section{Introduction}

One of the important developments in the field of ensemble-based models
\cite{Ferreira-Figueiredo-2012,Rokach-2010,Wozniak-etal-2014,ZH-Zhou-2012} of
the last years was a combination of several ensemble-based models, including
random forests (RFs) \cite{Breiman-1996} and the stacking, proposed by Zhou
and Feng \cite{Zhou-Feng-2017a} and called the Deep Forest (DF) or gcForest.
Its structure consists of layers similarly to a multi-layer neural network
structure, but each layer in gcForest contains many RFs instead of neurons.
gsForest can be regarded as an multi-layer ensemble of decision tree
ensembles. As pointed out by Zhou and Feng \cite{Zhou-Feng-2017a}, gcForest is
much easier to train and can perfectly work when there are only small-scale
training data in contrast to deep neural networks which require great effort
in hyperparameter tuning and large-scale training data. A lot of numerical
experiments provided by Zhou and Feng \cite{Zhou-Feng-2017a} illustrated that
gcForest outperforms many well-known methods or comparable with existing methods.

One of the motivations for the DF development is to build deep models based on
non-differentiable modules in contrast to deep neural networks which use
backpropagation required differentiability. Another important motivation is to
build deep models which require a small amount of training data due to a small
number of training parameters. As a result, the DF taking into the above has
exhibited high performance. This fact was a reason for developing new
modifications of the DF and for applying it to several applications. In
particular, Wang et al. \cite{Wang-Lu-etal-2018} proposed to apply the deep
forest to forecast the current health state or to diagnostic health
monitoring. Yang et al. \cite{Yang-Xu-Lian-Ji-2018} applied the DF to solving
the ship detection problem from thermal remote sensing imagery. Zheng et al.
\cite{Zheng-etal-2018} considered application of the DF to the pedestrian
detection problem in the framework of the extreme learning machine algorithms.
The application of the DF\ to sentiment analysis was given in
\cite{S-etal-2018}. Xia et al. \cite{Xia-Ming-Iwasaki-2018} used a DF
modification to fuse multiple sources remotely sensed datasets, such as
hyperspectral and Light Detection and Ranging (LiDAR)-derived digital surface
model, where ensembles of Rotation Forests and RFs are introduced. Zhao et al.
\cite{Zhao-etal-2018} proposed a deep forest-based protein location algorithm
relying on sequence information.

Guo et al. \cite{Guo-Liu-Li-Shang-2018} proposed a modification of the DF,
called BCDForest, to classify cancer subtypes on small-scale biology datasets.
The BCDForest uses a multi-class-grained scanning method to train multiple
binary classifiers and a boosting strategy to emphasize more important
features in cascade forests, thus to propagate the benefits of discriminative
features among cascade layers to improve the classification performance. The
same authors \cite{Guo-Liu-Li-Shang-2017} used the DF in order to solve the
same problem of the cancer subtype classification on gene expression data. Li
et al. \cite{Li-Zhang-Pan-Xie-Wu-Shi-2017} applied the DF to hyperspectral
image classification. Wen et al. \cite{Wen-Zhang-etal-2018} also proposed a
modification of the DF for developing efficient recommendation methods. The
modification is implemented by stacking Gradient Boosting Decision Trees in
the DF cascade structures.

Some improvements have been proposed by Utkin and Ryabinin
\cite{Utkin-Ryabinin-2018a,Utkin-Ryabinin-2018,Utkin-Ryabinin-2017a}. In
particular, modifications of the DF for solving the weakly supervised and
fully supervised metric learning problems were proposed in
\cite{Utkin-Ryabinin-2018} and \cite{Utkin-Ryabinin-2017a}, respectively. A
transfer learning model using the DF was presented in
\cite{Utkin-Ryabinin-2018a}. The main idea underlying the proposed
modifications is to assign weights to decision trees in every RF in order to
minimize the corresponding loss functions which depend on the problem solved.
The weights are used to replace the standard averaging of the class
probabilities for every instance and every decision tree with the weighted
average. Yin et al. \cite{Yin-Zhao-Yuan-Zhang-2018} introduced DF
reinforcement learning applied to large-scale interconnected power systems for
preventive strategy considering automatic generation control. Wu et al.
\cite{Wu-etal-2018} proposed an interesting modification of the DF, named the
multi-features fusion cascade XGBoost, for human facial age estimation by
means of a hierarchical regression approach. The modification uses the general
idea of the DF, but it consists of the cascade XGBoost models instead of RFs.
Another modification implemented and deployed the distributed version of the
DF model for automatic detection of a cash-out fraud was provided by Zhang et
al. \cite{Zhang-Zhou-etal-2018}. Han et al. \cite{Han-Li-Wan-Liu-2018}
proposed a combination of a convolution residual neural network with the DF
for scene recognition. In fact, the DF in this combination aims to classify an
extracted feature representation vector on the output of the residual neural network.

The same architecture of the cascade forest was proposed by Miller et al.
\cite{Miller-etal-2017}. However, this architecture differs from gcForest in
using only class vectors at the next cascade levels without their
concatenation with the original vector. Miller et al. \cite{Miller-etal-2017}
illustrated by numerical experiments that their approach is comparable to the
approach \cite{Zhou-Feng-2017a}. We have to point out that the cascade
structure with neural networks without backpropagation instead of forests was
proposed by Hettinger et al. \cite{Hettinger-etal-2017}.

One of the crucial shortcomings of the DF is that it passes all training and
testing instances through all levels of the cascade, leading to significant
increase of time complexity. In order to overcome this difficulty, another
improvement of the original DF was proposed by Pang et al.
\cite{Pang-etal-2018}, which significantly reduces the training and testing
times of forests at each level. According to the improvement, training
examples with high confidence (the maximum value of the estimated class
vector) directly pass to the final stage rather than passing through all the
levels. Pang et al. \cite{Pang-etal-2018} introduced a confidence screening
mechanism in the general framework of the DF, which categorizes instances at
every level of the cascade into two subsets: one is easy to predict; and the
other is hard. The improvement opens a door for developing new models
improving the DF.

Therefore, following the ideas of the DF improvement, we propose a new
modification of the confidence screening mechanism based on adaptive weighing
of every training instance at each cascade level depending on its mean class
vector at the previous level. It is called the Adaptive Weighted Deep Forest
(AWDF). Two ways are considered for applying weights. The first one is when
the weighted instances are randomly chosen for training trees in accordance
with their weights. This leads to reducing the set of \textquotedblleft
active\textquotedblright\ instances at every level of the forest cascade. The
second way is to use weights in implementing a splitting rule for training the
decision trees. The numerical experiments have shown that AWDF provides
outperformed results.

The paper is organized as follows. A short description of gcForest proposed by
Zhou and Feng \cite{Zhou-Feng-2017a} is given in Section 2. Section 3 provides
the confidence screening mechanism proposed by Pang et al.
\cite{Pang-etal-2018}. AWDF algorithm is considered in Section 4. Numerical
experiments with real data illustrating cases when the proposed AWDF
outperforms gcForest are given in Section 5. Concluding remarks are provided
in Section 6.

\section{A short introduction to Deep Forests}

\begin{figure}
[ptb]
\begin{center}
\includegraphics[
%natheight=4.139700in,
%natwidth=9.329200in,
height=1.8498in,
width=4.1537in
]%
{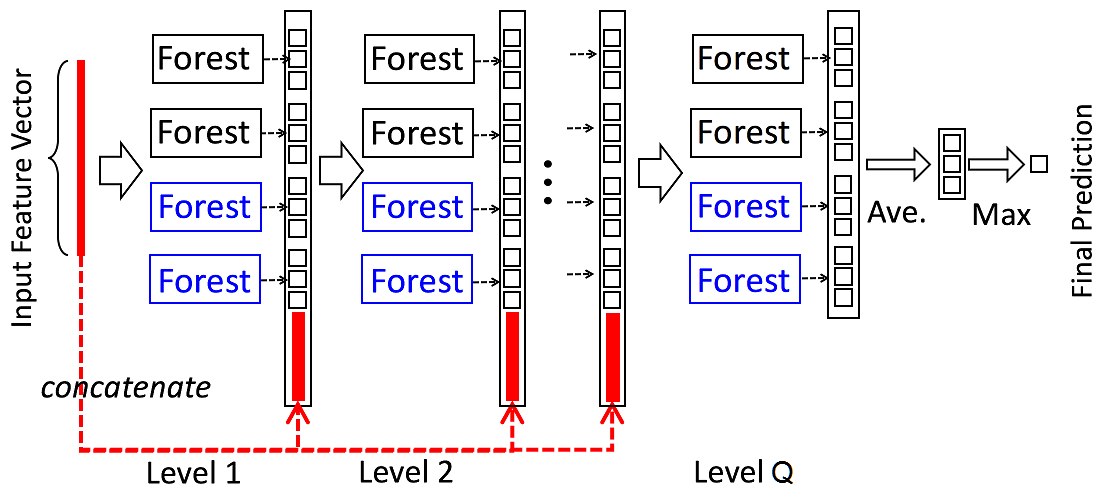}%
\caption{The architecture of the cascade forest \cite{Zhou-Feng-2017a}}%
\label{fig:cascade_forest}%
\end{center}
\end{figure}

One of the important peculiarities of gcForest is its cascade structure
proposed by Zhou and Feng \cite{Zhou-Feng-2017a}. Every cascade is represented
as an ensemble of decision tree forests. The cascade structure is a part of a
total gcForest structure. It implements the idea of representation learning by
means of the layer-by-layer processing of raw features. Each level of cascade
structure receives feature information processed by its preceding level, and
outputs its processing result to the next level. The architecture of the
cascade proposed by Zhou and Feng \cite{Zhou-Feng-2017a} is shown in Fig.
\ref{fig:cascade_forest}. It can be seen from the figure that each level of
the cascade consists of two different pairs of RFs which generate
3-dimensional class vectors concatenated each other and with the original
input. It should be noted that this structure of forests can be modified in
order to improve the gcForest for a certain application. After the last level,
we have the feature representation of the input feature vector, which can be
classified in order to get the final prediction. The gcForest representational
learning ability is enhanced by applying the second part of gcForest called as
the so-called multi-grained scanning. The multi-grained scanning structure
uses sliding windows to scan the raw features. Its output is a set of feature
vectors produced by sliding windows of multiple sizes. We mainly pay attention
to the first part of gcForest because our modification relates to the RFs.

\begin{figure}
[ptb]
\begin{center}
\includegraphics[
%natheight=2.219100in,
%natwidth=3.255200in,
height=2.2589in,
width=3.3001in
]%
{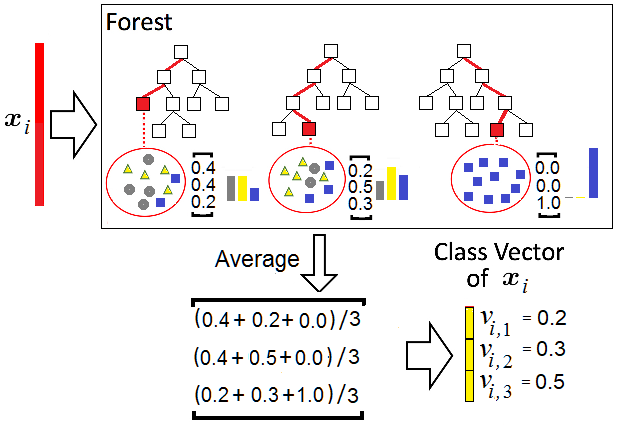}%
\caption{An illustration of the class vector generation by using average of
the tree probability class vectors}%
\label{fig:weighted_class_vector_gen_7_0}%
\end{center}
\end{figure}

Given an instance, each forest produces an estimate of a class distribution by
counting the percentage of different classes of examples at the leaf node
where the concerned instance falls into, and then averaging across all trees
in the same forest as it is schematically shown in Fig.
\ref{fig:weighted_class_vector_gen_7_0}. The class distribution forms a class
vector, which is then concatenated with the original vector to be input to the
next level of cascade. The usage of the class vector as a result of the RF
classification is very similar to the idea underlying the stacking algorithm
\cite{Wolpert-1992} which trains the first-level learners using the original
training dataset. Then the stacking algorithm generates a new dataset for
training the second-level learner (meta-learner) such that the outputs of the
first-level learners are regarded as input features for the second-level
learner while the original labels are still regarded as labels of the new
training data. In contrast to the standard stacking algorithm, gcForest
simultaneously uses the original vector and the class vectors (meta-learners)
at the next level of cascade by means of their concatenation. This implies
that the feature vector is enlarged and enlarged after every cascade level.
After the last level, we have the feature representation of the input feature
vector, which can be classified in order to get the final prediction. Zhou and
Feng \cite{Zhou-Feng-2017a} propose to use different forests at every level in
order to provide the diversity which is an important requirement for the RF construction.%

%TCIMACRO{\FRAME{ftbpFU}{4.1537in}{1.8498in}{0pt}{\Qcb{The architecture of the
%cascade forest \cite{Zhou-Feng-2017a}}}{\Qlb{fig:cascade_forest}%
%}{forest_cascade.png}{\special{ language "Scientific Word";  type "GRAPHIC";
%maintain-aspect-ratio TRUE;  display "USEDEF";  valid_file "F";
%width 4.1537in;  height 1.8498in;  depth 0pt;  original-width 9.3292in;
%original-height 4.1397in;  cropleft "0";  croptop "1";  cropright "1";
%cropbottom "0";  filename 'Forest_Cascade.png';file-properties "XNPEU";}} }%
%BeginExpansion
%EndExpansion
%

%TCIMACRO{\FRAME{ftbpFU}{3.3001in}{2.2589in}{0pt}{\Qcb{An illustration of the
%class vector generation by using average of the tree probability class
%vectors}}{\Qlb{fig:weighted_class_vector_gen_7_0}}%
%{weighted_class_vector_gen_7_2.png}{\special{ language "Scientific Word";
%type "GRAPHIC";  maintain-aspect-ratio TRUE;  display "USEDEF";
%valid_file "F";  width 3.3001in;  height 2.2589in;  depth 0pt;
%original-width 3.2552in;  original-height 2.2191in;  cropleft "0";
%croptop "1";  cropright "1";  cropbottom "0";
%filename 'weighted_class_vector_gen_7_2.png';file-properties "XNPEU";}} }%
%BeginExpansion
%EndExpansion

\section{The confidence screening mechanism}

According to \cite{Pang-etal-2018}, the main idea underlying the confidence
screening mechanism is that an instance is pushed to the next level of the
cascade only if it is determined to require a higher level of learning;
otherwise, it is predicted using the model at the current level.

We will consider the standard classification problem which can be formally
written as follows. Given $n$ training instances $S=\{(\mathbf{x}_{1}%
,y_{1}),(\mathbf{x}_{2},y_{2}),...,(\mathbf{x}_{n},y_{n})\}$, in which
$\mathbf{x}_{i}\in\mathbb{R}^{m}$ represents a feature vector involving $m$
features and $y_{i}\in\{1,...,C\}$ represents the class of the associated
instances, the task of classification is to construct an accurate classifier
$c:$ $\mathbb{R}^{m}\rightarrow\{1,...,C\}$ that maximizes the probability
that $c(\mathbf{x}_{i})=y_{i}$ for $i=1,...,n$.

A decision tree in every forest produces an estimate of the class probability
distribution $\mathbf{p}=(p_{1},...,p_{C})$ by counting the percentage of
different classes of training instances at the leaf node where the concerned
instance falls into. Then the class probabilities $\mathbf{v}_{i}=\left(
v_{i,1},...,v_{i,C}\right)  $ of $\mathbf{x}_{i}$ for every RF are computed by
averaging all class probability distributions $p$ across all trees as it is
shown in Fig. \ref{fig:weighted_class_vector_gen_7_0}, where we partly modify
a picture from \cite{Zhou-Feng-2017a} in order to illustrate how elements of
the class vector are derived as a simple sum.

Suppose that all RFs have the same number $T$ of decision trees, every cascade
level contains $M$ RFs, and the number of cascade levels is $Q$. Then a
current level of the cascade produces $M$ class vectors $\mathbf{v}%
_{i,1},...,\mathbf{v}_{i,M}$ which are then concatenated with the original
vector $\mathbf{x}_{i}$ to be input to the next level of the cascade, i.e.,
the training set for the next level is defined as
\[
S^{\ast}=\{((\mathbf{x}_{i},\mathbf{v}_{i,1},...,\mathbf{v}_{i,M}%
),y_{i}),i=1,...,n\}.
\]

At level $q$, if the prediction confidence of one instance is larger than
threshold $\eta_{q}$, then its final prediction is produced at the current
level, otherwise it needs to go through the next level (and potentially all
levels in the cascade). One of the ways to define the prediction confidence of
instance $((\mathbf{x}_{i},\mathbf{v}_{i,1},...,\mathbf{v}_{i,M}),y_{i})$ is
to find the mean vector of class probabilities, namely,
\[
\mathbf{v}_{i}=\frac{1}{M}\sum_{k=1}^{M}\mathbf{v}_{i,k}.
\]

Let us introduce the indicator $I$ defined as
\[
I=\left\{
\begin{array}
[c]{cc}%
1, & \max(v_{i,1},...,v_{i,C})\geq\eta_{q},\\
0, & \text{otherwise.}%
\end{array}
\right.
\]

The choice of the threshold $\eta_{q}$ is considered by Pang et al.
\cite{Pang-etal-2018} in detail, where $\eta_{q}$ at level $q$ is determined
automatically based on the cross-validated error rate of all the training instances.

If the indicator $I$ is $0$, then the feature vector $\mathbf{x}_{i}$ has to
go through the next level. If $I=1$, the final prediction is produced at the
current level during testing such that
\[
y_{i}=\arg_{c=1,...,C}\max v_{i,c}.
\]
During training, we do not need to go through the next level if $I=1$ and
\[
\arg_{c=1,...,C}\max v_{i,c}=y_{i},
\]
otherwise the instance has to go through the next level $q+1$.

\section{The adaptive weighted deep forest}

We propose to assign a weight to every instance $\mathbf{x}_{i}$ at a current
forest cascade level in accordance with its mean class vector $\mathbf{v}_{i}$
at the previous level. Let us introduce the vector $\mathbf{o}_{i}%
=(0,...,0,1_{y_{i}},0,...,0)$, where the index of the unit element is $y_{i}$.
Then the weight $w_{i}$ is determined as a function $f$ of a distance between
the mean vector of class probabilities $\mathbf{v}_{i}$ and the vector
$\mathbf{o}_{i}$, denoted as $d(\mathbf{v}_{i},\mathbf{o}_{i})$. Suppose that
we get the mean class vector $\mathbf{v}_{i}$ for instance $\mathbf{x}_{i}$ at
the current level $q$. Then we can write the weight of instance $\mathbf{x}%
_{i}$ as
\[
w_{i}=f(d(\mathbf{v}_{i},\mathbf{o}_{i})).
\]

The weight $w_{i}$ is used for training RFs at the next level $q+1$. It is
obvious that the function $f$ increases with $d(\mathbf{v}_{i},\mathbf{o}%
_{i})$. In particular, if instance $\mathbf{x}_{i}$ is correctly classified at
level $q$ such that the distance $d(\mathbf{v}_{i},\mathbf{o}_{i})$ is $0$,
then the weight $w_{i}$ has to be $0$. In this case, the instance is not used
at the next level. In other words, due to the small weight, the instance
$\mathbf{x}_{i}$ will have lesser chance to appear in the trees of the next
level compared to other instances. If the distance is $1$ ($\mathbf{x}_{i}$ is
totally incorrectly classified), then the weight $w_{i}$ has to be also $1$ or
to have some maximal value. Simple examples of the function $f$ are
$w_{i}=\left(  d(\mathbf{v}_{i},\mathbf{o}_{i})\right)  ^{2}$ or
$w_{i}=d(\mathbf{v}_{i},\mathbf{o}_{i})$. Moreover, the distance can be also
differently taken. One of the most popular distances is Euclidean one, i.e.,
$d(\mathbf{v}_{i},\mathbf{o}_{i})=||\mathbf{v}_{i}-\mathbf{o}_{i}||_{2}$.

Another way for determining the weights is to consider the following
function:
\[
w_{i}=1-f(\mathbf{v}_{i}\cdot\mathbf{o}_{i}^{\mathrm{T}}).
\]

In this case, we analyze only a probability of the class $y_{i}$, i.e.,
$v_{i,y_{i}}$. If this probability of an instance is close to $1$ (correct
classification), then the corresponding weight of the instance is close to
$0$. If the probability of the instance is close to $0$ (incorrect
classification), then the corresponding weight is close to $1$. A simplest
case is $w_{i}=1-\mathbf{v}_{i}\cdot\mathbf{o}_{i}^{\mathrm{T}}$. This
definition of weights almost coincides with the rule for decision about going
the instance through the next level in the confidence screening mechanism (see
the previous section).

It should be noted that the normalized weights $w_{1},...,w_{n}$ define a
probability distribution on instances in the training data, which can be used
for building decision trees of a RF.

We define two strategies for using the weights. In accordance with the first
strategy, we randomly draw instances from the training set with replacement in
accordance with this probability distribution. If the weight of the $i$-th
instance is very close to $0$, it does not take part in building decision
trees. In other words, instances with a high prediction confidence are
predicted using the model at the current level.

In addition to the above weighted procedure, we can also introduce a threshold
$\eta_{q}$ to compare it with the value $1-w_{i}$. If the $1-w_{i}\geq\eta
_{q}$, then the corresponding instance is predicted by using the model at
current level. In sum, the number of instances for training are reduced at
every level simplifying the whole training process. However, if the training
set is imbalanced, then the number of randomly drawn instances of a class may
be very small to be used for training.

The proposed approach is intuitively very close to the AdaBoost algorithm
\cite{Freund-Shapire-97}, where instances from a training set are drawn for
classification from an iteratively updated sample distribution defined on
elements of the training set. Every level of the DF can be viewed as an
iteration in AdaBoost. The sample distribution ensures that instances
misclassified by the previous classifier (at the previous iteration) are more
likely to be included in the training data of the next classifier. In each
iteration, the weights of all misclassified instances are increased while the
weights of correctly classified examples are decreased.

According to the second strategy, we apply the procedure of direct use of the
weights during learning in a splitting rule, which is implemented in many
versions of the decision tree algorithms, for example, in C4.5 and CART. In
particular, the weights and the entropy measure are combined in the splitting
rule. The weights are again viewed as probabilities of instances and used in
definition of the entropy measure. This strategy does not simplify the whole
deep forest training process because all instances are used for training the
trees. However, the problem of a lack of instances of a certain class at some
level is avoided in this case.

\section{Numerical experiments}

In order to illustrate AWDF, we investigate the model for datasets from UCI
Machine Learning Repository \cite{Dua:2017}. Table \ref{t:IDF_datasets} is a
brief introduction about these data sets, while more detailed information can
be found from, respectively, the data resources. Table \ref{t:IDF_datasets}
shows the number of features $m$ for the corresponding dataset, the number of
training instances $n$ and the number of classes $C$. We also use the IMDB
dataset \cite{Mass-etal-2011} for the sentiment classification, which consists
of 25000 movie reviews for training and 25000 for testing. The reviews are
represented by tf-idf features. The IMDB dataset is available at
http://ai.stanford.edu/\symbol{126}amaas/data/sentiment.

AWDF uses a software in Python which implements gcForest and is available at
http://lamda.nju.edu.cn/code\_gcForest.ashx. AWDF has the same cascade
structure as the standard gcForest described in \cite{Zhou-Feng-2017a} (two
completely-random tree forests and two random forests at every level,
completely-random trees are generated by randomly selecting a feature for
split at each node of the tree). The forest cascade is used for numerical
experiments without the Multi-Grained Scanning part of gcForest. Accuracy
measure $A$ used in numerical experiments is the proportion of correctly
classified cases on a sample of data. To evaluate the average accuracy, we
perform a cross-validation with $50$ repetitions, where in each run, we
randomly select $n_{\text{tr}}=4n/5$ training data and $n_{\text{test}}=n/5$
testing data. We apply the second strategy to training RFs by using the
weighted instances because we are interesting in increasing the accuracy of
AWDF, but not training or testing time. However, we also provide results for
the first strategy. %

%TCIMACRO{\TeXButton{B}{\begin{table}[tbp] \centering}}%
%BeginExpansion
\begin{table}[tbp] \centering
%EndExpansion
\caption{A brief introduction about data sets}%
\begin{tabular}
[c]{ccccc}\hline
Data set & Abbreviation & $m$ & $n$ & $C$\\\hline
Adult Income & Adult & $14$ & $48842$ & $2$\\\hline
Car & Car & $6$ & $1728$ & $4$\\\hline
Diabetic Retinopathy & Diabet & $20$ & $1151$ & $2$\\\hline
EEG Eye State & EEG & $15$ & $14980$ & $2$\\\hline
Haberman's Breast Cancer Survival & Haberman & $3$ & $306$ & $2$\\\hline
Ionosphere & Ion & $34$ & $351$ & $2$\\\hline
Seeds & Seeds & $7$ & $210$ & $3$\\\hline
Seismic Mining & Seismic & $19$ & $2584$ & $2$\\\hline
Teaching Assistant Evaluation & TAE & $5$ & $151$ & $3$\\\hline
Tic-Tac-Toe Endgame & TTTE & $9$ & $958$ & $2$\\\hline
Website Phishing & Website & $30$ & $2456$ & $3$\\\hline
Wholesale Customer Region & WCR & $8$ & $440$ & $3$\\\hline
Letter & Letter & $16$ & $20000$ & $26$\\\hline
Yeast & Yeast & $8$ & $1484$ & $10$\\\hline
Nursery & Nursery & $8$ & $12960$ & $5$\\\hline
Ecoli & Ecoli & $8$ & $336$ & $8$\\\hline
Dermatology & Dermatology & $33$ & $366$ & $6$\\\hline
IMDB & IMDB & $5000$ & $50000$ & $2$\\\hline
\end{tabular}
\label{t:IDF_datasets}%
%TCIMACRO{\TeXButton{E}{\end{table}}}%
%BeginExpansion
\end{table}%
%EndExpansion

First of all, our aim is to compare AWDF with gcForest and to consider
different cases of the weight function $f$. We denote $w_{i}=1-(\mathbf{v}%
_{i}\cdot\mathbf{o}_{i}^{\mathrm{T}})^{2}$ as $1-w^{2}$; $w_{i}=1-(\mathbf{v}%
_{i}\cdot\mathbf{o}_{i}^{\mathrm{T}})^{1/2}$ as $1-w^{1/2}$; $w_{i}%
=1-(\mathbf{v}_{i}\cdot\mathbf{o}_{i}^{\mathrm{T}})$ as $1-w$; $w_{i}%
=||\mathbf{v}_{i}-\mathbf{o}_{i}||_{2}$ (Euclidean distance) as $L_{2}$.
Numerical results of comparison of gcForest (gcF) and four cases of the weight
definition are shown in Table \ref{t:AWDF1}, where the first column contains
abbreviations of the tested data sets, the second column is the accuracy
measure by using gcForest, other columns correspond to the accuracy measures
of AWDF by different functions of weights. It should be noted that the largest
values of the accuracy measures obtained by different numbers of trees are
shown in Table \ref{t:AWDF1}. It can be seen from Table \ref{t:AWDF1} that at
least one of the cases of the proposed AWDF outperforms gcForest for most
considered data sets. The best performance on each dataset is shown in bold.
We note that our method yields the best accuracy on 16 out of 18 datasets tested.

We can also conclude from Table \ref{t:AWDF1} that there is no the best choice
of the weight function for all datasets. Though, we can also see that the
function $w_{i}=1-(\mathbf{v}_{i}\cdot\mathbf{o}_{i}^{\mathrm{T}})^{1/2}$
provides the largest number of the best results. This function makes weights
to be close to $0$ if an instance is correctly classified. At the same time,
the \textquotedblleft bad\textquotedblright\ instances have weights close to
$1$, and they introduce a large impact in computing the splitting rule. In
contrast to this function, function $1-(\mathbf{v}_{i}\cdot\mathbf{o}%
_{i}^{\mathrm{T}})^{2}$ shows worse results. This is due to the fact that the
resulting weights are close to $0.5$. As a result, the difference between
weights of correctly and incorrectly classified instances is smaller, and the
separation effect is reduced.%

%TCIMACRO{\TeXButton{B}{\begin{table}[tbp] \centering}}%
%BeginExpansion
\begin{table}[tbp] \centering
%EndExpansion
\caption{Accuracy measures gcForest and four cases of the weight function for the second strategy of using weights}%
\begin{tabular}
[c]{cccccc}\hline
Data set & gcF & $1-w^{2}$ & $1-w^{1/2}$ & $L_{2}$ & $1-w$\\\hline
Adult & $86.12$ & $86.20$ & $86.26$ & $86.11$ & $\mathbf{86.30}$\\\hline
Car & $98.28$ & $98.42$ & $\mathbf{98.74}$ & $98.45$ & $98.51$\\\hline
Diabet & $69.05$ & $69.06$ & $\mathbf{69.69}$ & $69.16$ & $69.13$\\\hline
EEG & $95.75$ & $95.82$ & $\mathbf{96.16}$ & $95.74$ & $95.94$\\\hline
Haberman & $74.15$ & $73.90$ & $\mathbf{74.77}$ & $74.09$ & $73.65$\\\hline
Ion & $94.32$ & $94.32$ & $94.37$ & $94.26$ & $\mathbf{94.64}$\\\hline
Seeds & $93.17$ & $\mathbf{93.62}$ & $\mathbf{93.62}$ & $93.08$ &
$92.72$\\\hline
Seismic & $93.50$ & $93.63$ & $93.53$ & $\mathbf{93.67}$ & $93.61$\\\hline
TAE & $52.76$ & $53.76$ & $55.14$ & $55.26$ & $\mathbf{55.76}$\\\hline
TTTE & $99.03$ & $99.09$ & $\mathbf{99.11}$ & $\mathbf{99.11}$ &
$98.99$\\\hline
Website & $90.15$ & $90.45$ & $\mathbf{90.59}$ & $90.00$ & $90.25$\\\hline
WCR & $\mathbf{72.81}$ & $72.47$ & $72.64$ & $72.29$ & $72.47$\\\hline
Letter & $97.14$ & $97.06$ & $97.08$ & $\mathbf{97.18}$ & $97.07$\\\hline
Yeast & $63.03$ & $\mathbf{62.92}$ & $63.26$ & $62.56$ & $62.81$\\\hline
Nursery & $66.40$ & $66.49$ & $66.56$ & $\mathbf{66.67}$ & $66.28$\\\hline
Ecoli & $\mathbf{89.15}$ & $85.27$ & $87.50$ & $89.14$ & $84.52$\\\hline
Dermatology & $58.10$ & $59.47$ & $59.47$ & $59.21$ & $\mathbf{60.30}$\\\hline
IMDB & $88.74$ & $89.21$ & $\mathbf{89.44}$ & $88.95$ & $89.35$\\\hline
\end{tabular}
\label{t:AWDF1}%
%TCIMACRO{\TeXButton{E}{\end{table}}}%
%BeginExpansion
\end{table}%
%EndExpansion

In order to formally show the outperformance of the proposed AWDF, we apply
the $t$-test which has been proposed and described by Demsar
\cite{Demsar-2006} for testing whether the average difference in the
performance of two classifiers AWDF and gcForest is significantly different
from zero. We can use the differences between accuracy measures of AWDF and
gcForest, and then to compare them with $0$. The $t$ statistics in this case
is distributed according to the Student distribution with $18-1$ degrees of
freedom. The results of computing the $t$ statistics for the difference
between the best value of AWDF and gcForest are the p-values denoted as $p$
and the $95\%$ confidence interval for the mean $0.557$, which are $p=0.0083$
and $[0.164,0.951]$, respectively. The $t$-test demonstrates the clear
outperforming of the proposed model in comparison with gcForest.

\begin{figure}
[ptb]
\begin{center}
\includegraphics[
%natheight=5.000400in,
%natwidth=13.765200in,
height=1.9095in,
width=5.2399in
]%
{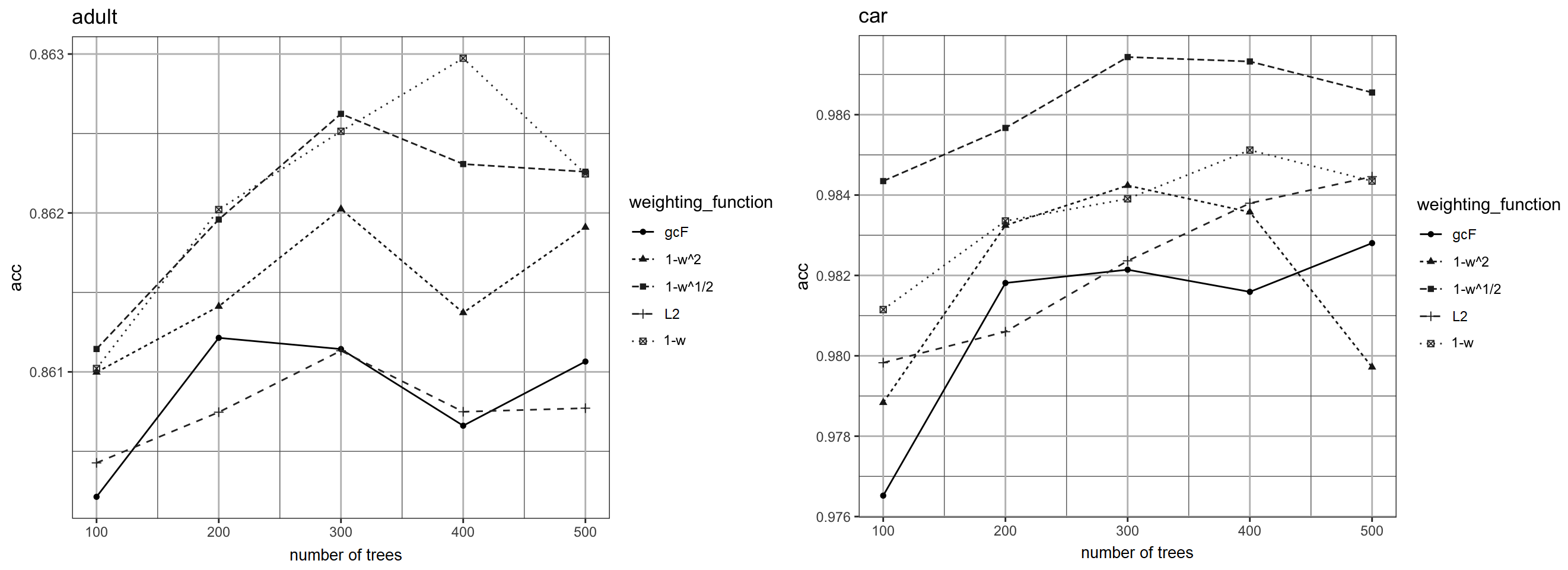}%
\caption{Accuracy measures as a function of the number of trees for the Adult
and Car datasets}%
\label{fig:as_car}%
\end{center}
\end{figure}

\begin{figure}
[ptb]
\begin{center}
\includegraphics[
%natheight=5.000400in,
%natwidth=13.749600in,
height=1.9207in,
width=5.2676in
]%
{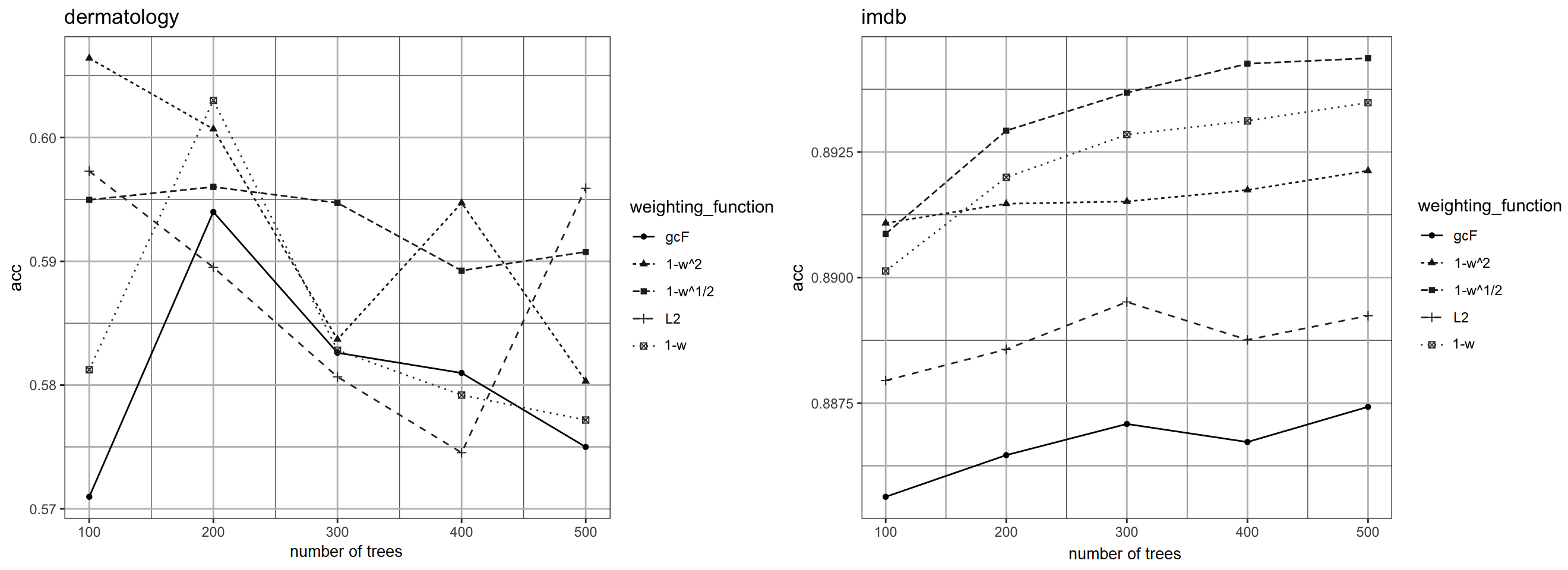}%
\caption{Accuracy measures as a function of the number of trees for the
Dermatology and IMDB datasets}%
\label{fig:dermat_IMDB}%
\end{center}
\end{figure}

In order to investigate how the accuracy measures depend on the number of
decision trees in the RF, we depict the corresponding dependencies in Figs.
\ref{fig:as_car}-\ref{fig:dermat_IMDB}. We compare gcForest and AWDF by four
different weight functions. We again can see from Figs. \ref{fig:as_car}%
-\ref{fig:dermat_IMDB} that AWDF outperforms gcForest for all datasets. Table
\ref{t:AWDF1} is composed from largest values of accuracy measures given in
Figs. \ref{fig:as_car}-\ref{fig:dermat_IMDB}.%

%TCIMACRO{\FRAME{ftbpFU}{5.2399in}{1.9095in}{0pt}{\Qcb{Accuracy measures as a
%function of the number of trees for the Adult and Car datasets}}%
%{\Qlb{fig:as_car}}{adult-car.png}{\special{ language "Scientific Word";
%type "GRAPHIC";  maintain-aspect-ratio TRUE;  display "USEDEF";
%valid_file "F";  width 5.2399in;  height 1.9095in;  depth 0pt;
%original-width 13.7652in;  original-height 5.0004in;  cropleft "0";
%croptop "1";  cropright "1";  cropbottom "0";
%filename 'Adult-Car.png';file-properties "XNPEU";}} }%
%BeginExpansion
%EndExpansion
%

%TCIMACRO{\FRAME{ftbpFU}{5.2814in}{1.9147in}{0pt}{\Qcb{Accuracy measures as a
%function of the number of trees for the Diabetic and EEG datasets}}%
%{}{diabetic-eeg.png}{\special{ language "Scientific Word";  type "GRAPHIC";
%maintain-aspect-ratio TRUE;  display "USEDEF";  valid_file "F";
%width 5.2814in;  height 1.9147in;  depth 0pt;  original-width 13.8335in;
%original-height 5.0004in;  cropleft "0";  croptop "1";  cropright "1";
%cropbottom "0";  filename 'Diabetic-EEG.png';file-properties "XNPEU";}} }%
%BeginExpansion
\begin{figure}
[ptb]
\begin{center}
\includegraphics[
%natheight=5.000400in,
%natwidth=13.833500in,
height=1.9147in,
width=5.2814in
]%
{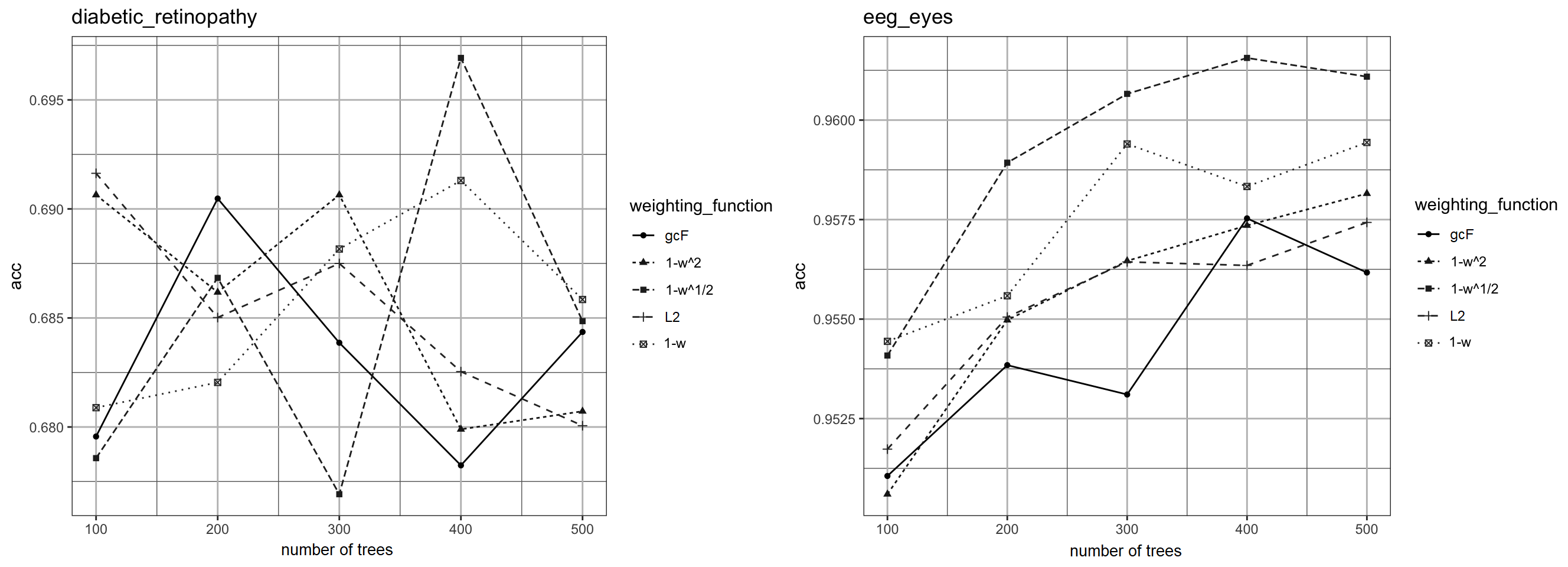}%
\caption{Accuracy measures as a function of the number of trees for the
Diabetic and EEG datasets}%
\end{center}
\end{figure}
%EndExpansion
%

%TCIMACRO{\FRAME{ftbpFU}{5.2952in}{1.9311in}{0pt}{\Qcb{Accuracy measures as a
%function of the number of trees for the Haberman and Ionosphere datasets}}%
%{}{haberman-ion.png}{\special{ language "Scientific Word";  type "GRAPHIC";
%maintain-aspect-ratio TRUE;  display "USEDEF";  valid_file "F";
%width 5.2952in;  height 1.9311in;  depth 0pt;  original-width 13.76in;
%original-height 5.0004in;  cropleft "0";  croptop "1";  cropright "1";
%cropbottom "0";  filename 'Haberman-Ion.png';file-properties "XNPEU";}} }%
%BeginExpansion
\begin{figure}
[ptb]
\begin{center}
\includegraphics[
%natheight=5.000400in,
%natwidth=13.760000in,
height=1.9311in,
width=5.2952in
]%
{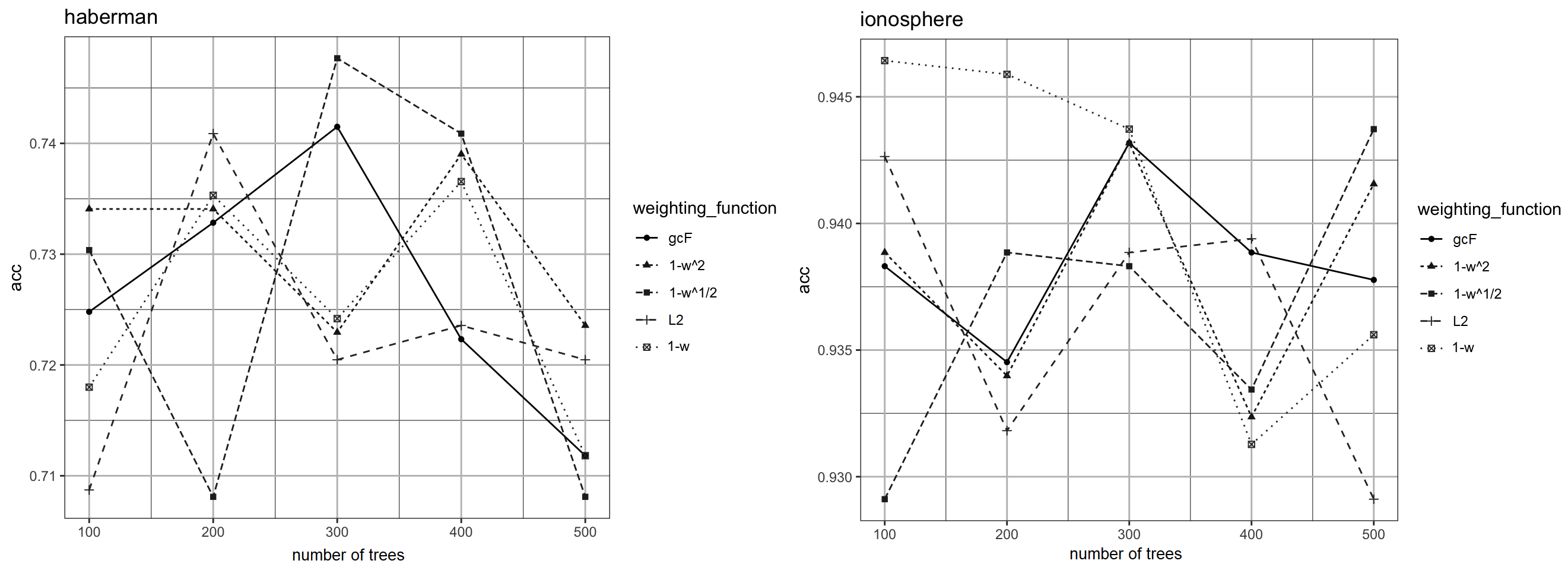}%
\caption{Accuracy measures as a function of the number of trees for the
Haberman and Ionosphere datasets}%
\end{center}
\end{figure}
%EndExpansion
%

%TCIMACRO{\FRAME{ftbpFU}{5.214in}{1.8965in}{0pt}{\Qcb{Accuracy measures as a
%function of the number of trees for the Seeds and Seismic datasets}}%
%{}{seeds-seismic.png}{\special{ language "Scientific Word";  type "GRAPHIC";
%maintain-aspect-ratio TRUE;  display "USEDEF";  valid_file "F";
%width 5.214in;  height 1.8965in;  depth 0pt;  original-width 13.792in;
%original-height 5.0004in;  cropleft "0";  croptop "1";  cropright "1";
%cropbottom "0";  filename 'Seeds-Seismic.png';file-properties "XNPEU";}} }%
%BeginExpansion
\begin{figure}
[ptb]
\begin{center}
\includegraphics[
%natheight=5.000400in,
%natwidth=13.792000in,
height=1.8965in,
width=5.214in
]%
{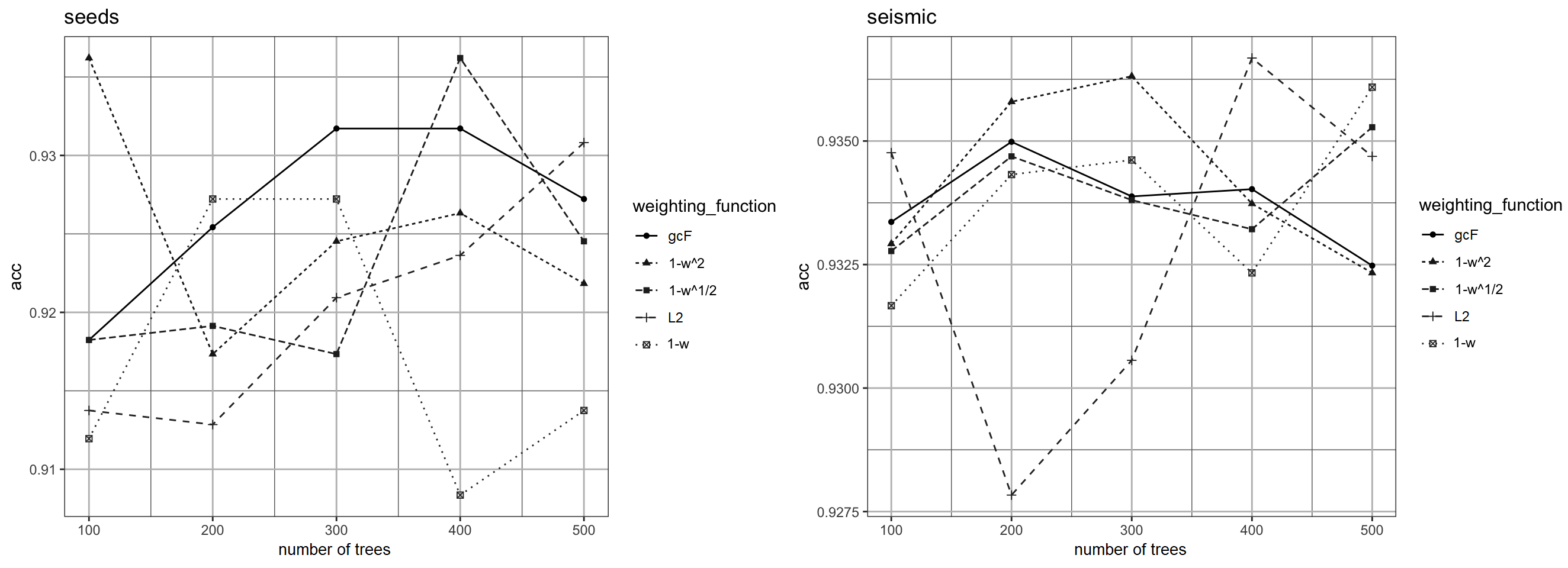}%
\caption{Accuracy measures as a function of the number of trees for the Seeds
and Seismic datasets}%
\end{center}
\end{figure}
%EndExpansion
%

%TCIMACRO{\FRAME{ftbpFU}{5.2226in}{1.9061in}{0pt}{\Qcb{Accuracy measures as a
%function of the number of trees for the TAE and TTTE datasets}}{}%
%{teaching_ass_tic_tac_toe.png}{\special{ language "Scientific Word";
%type "GRAPHIC";  maintain-aspect-ratio TRUE;  display "USEDEF";
%valid_file "F";  width 5.2226in;  height 1.9061in;  depth 0pt;
%original-width 13.7393in;  original-height 5.0004in;  cropleft "0";
%croptop "1";  cropright "1";  cropbottom "0";
%filename 'Teaching_ass_Tic_tac_toe.png';file-properties "XNPEU";}} }%
%BeginExpansion
\begin{figure}
[ptb]
\begin{center}
\includegraphics[
%natheight=5.000400in,
%natwidth=13.739300in,
height=1.9061in,
width=5.2226in
]%
{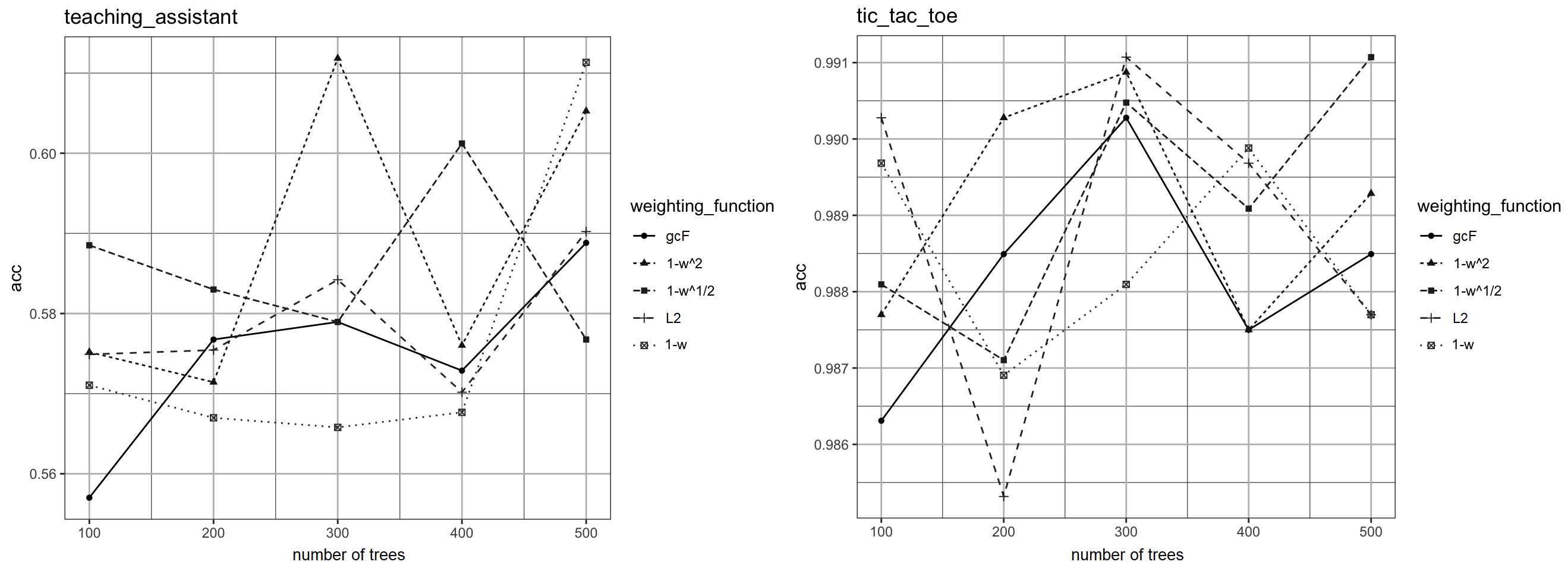}%
\caption{Accuracy measures as a function of the number of trees for the TAE
and TTTE datasets}%
\end{center}
\end{figure}
%EndExpansion
%

%TCIMACRO{\FRAME{ftbpFU}{5.2503in}{1.9086in}{0pt}{\Qcb{Accuracy measures as a
%function of the number of trees for the Website and WCR datasets}}%
%{}{website-wcr.png}{\special{ language "Scientific Word";  type "GRAPHIC";
%maintain-aspect-ratio TRUE;  display "USEDEF";  valid_file "F";
%width 5.2503in;  height 1.9086in;  depth 0pt;  original-width 13.8024in;
%original-height 5.0004in;  cropleft "0";  croptop "1";  cropright "1";
%cropbottom "0";  filename 'Website-WCR.png';file-properties "XNPEU";}} }%
%BeginExpansion
\begin{figure}
[ptb]
\begin{center}
\includegraphics[
%natheight=5.000400in,
%natwidth=13.802400in,
height=1.9086in,
width=5.2503in
]%
{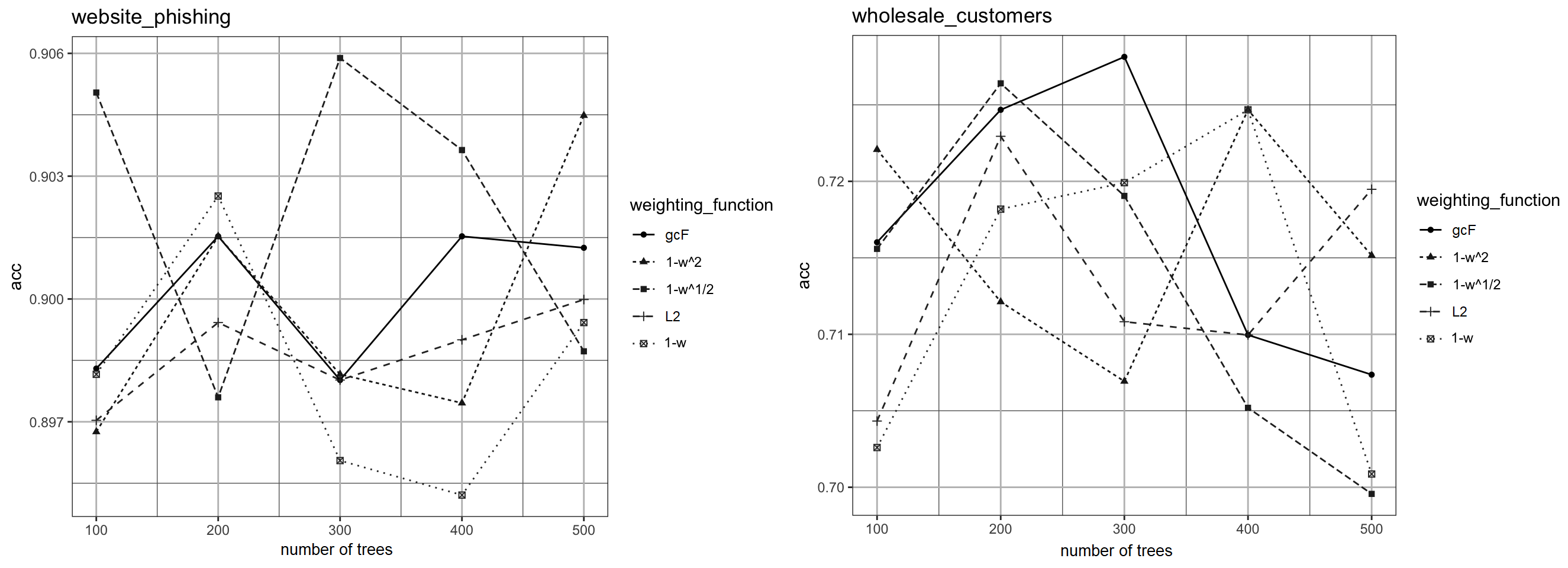}%
\caption{Accuracy measures as a function of the number of trees for the
Website and WCR datasets}%
\end{center}
\end{figure}
%EndExpansion
%

%TCIMACRO{\FRAME{ftbpFU}{5.2555in}{1.913in}{0pt}{\Qcb{Accuracy measures as a
%function of the number of trees for the Letter and Yeast datasets}}%
%{}{letter-yeast.png}{\special{ language "Scientific Word";  type "GRAPHIC";
%maintain-aspect-ratio TRUE;  display "USEDEF";  valid_file "F";
%width 5.2555in;  height 1.913in;  depth 0pt;  original-width 13.7816in;
%original-height 5.0004in;  cropleft "0";  croptop "1";  cropright "1";
%cropbottom "0";  filename 'Letter-Yeast.png';file-properties "XNPEU";}} }%
%BeginExpansion
\begin{figure}
[ptb]
\begin{center}
\includegraphics[
%natheight=5.000400in,
%natwidth=13.781600in,
height=1.913in,
width=5.2555in
]%
{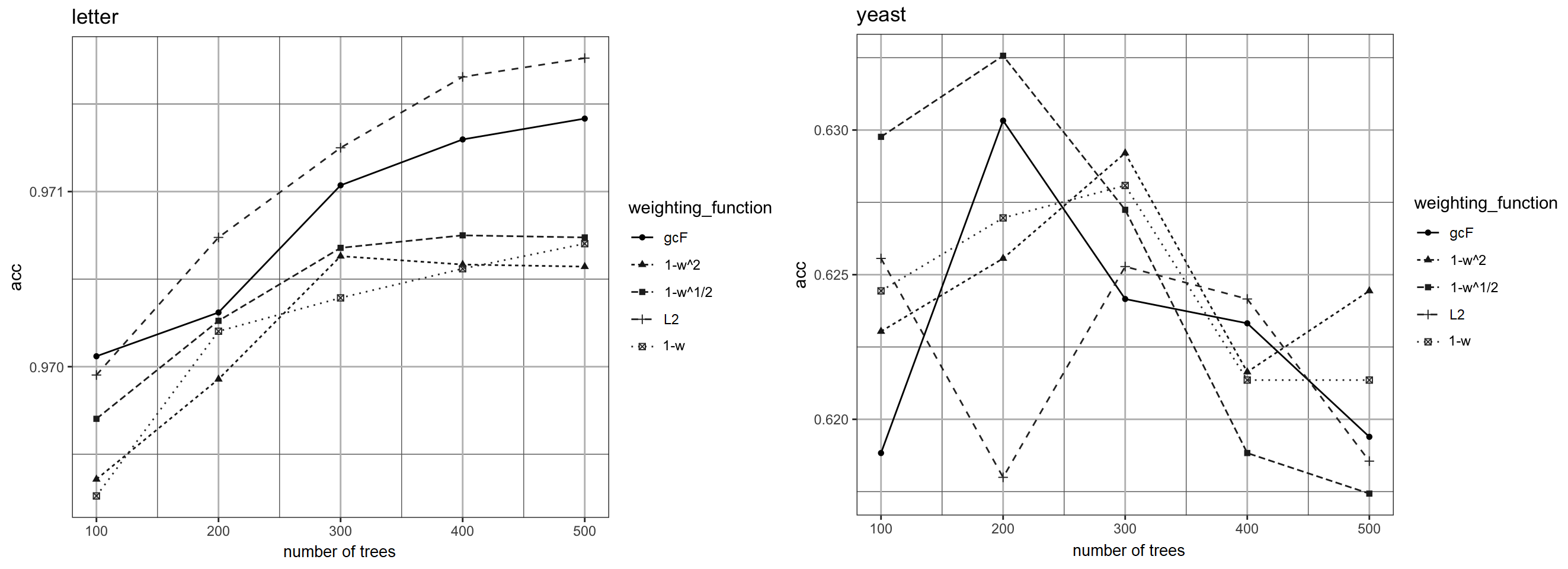}%
\caption{Accuracy measures as a function of the number of trees for the Letter
and Yeast datasets}%
\end{center}
\end{figure}
%EndExpansion
%

%TCIMACRO{\FRAME{ftbpFU}{5.2615in}{1.9035in}{0pt}{\Qcb{Accuracy measures as a
%function of the number of trees for the Nursery and Ecoli datasets}}%
%{}{nursery-ecoli.png}{\special{ language "Scientific Word";  type "GRAPHIC";
%maintain-aspect-ratio TRUE;  display "USEDEF";  valid_file "F";
%width 5.2615in;  height 1.9035in;  depth 0pt;  original-width 13.8647in;
%original-height 5.0004in;  cropleft "0";  croptop "1";  cropright "1";
%cropbottom "0";  filename 'Nursery-Ecoli.png';file-properties "XNPEU";}} }%
%BeginExpansion
\begin{figure}
[ptb]
\begin{center}
\includegraphics[
%natheight=5.000400in,
%natwidth=13.864700in,
height=1.9035in,
width=5.2615in
]%
{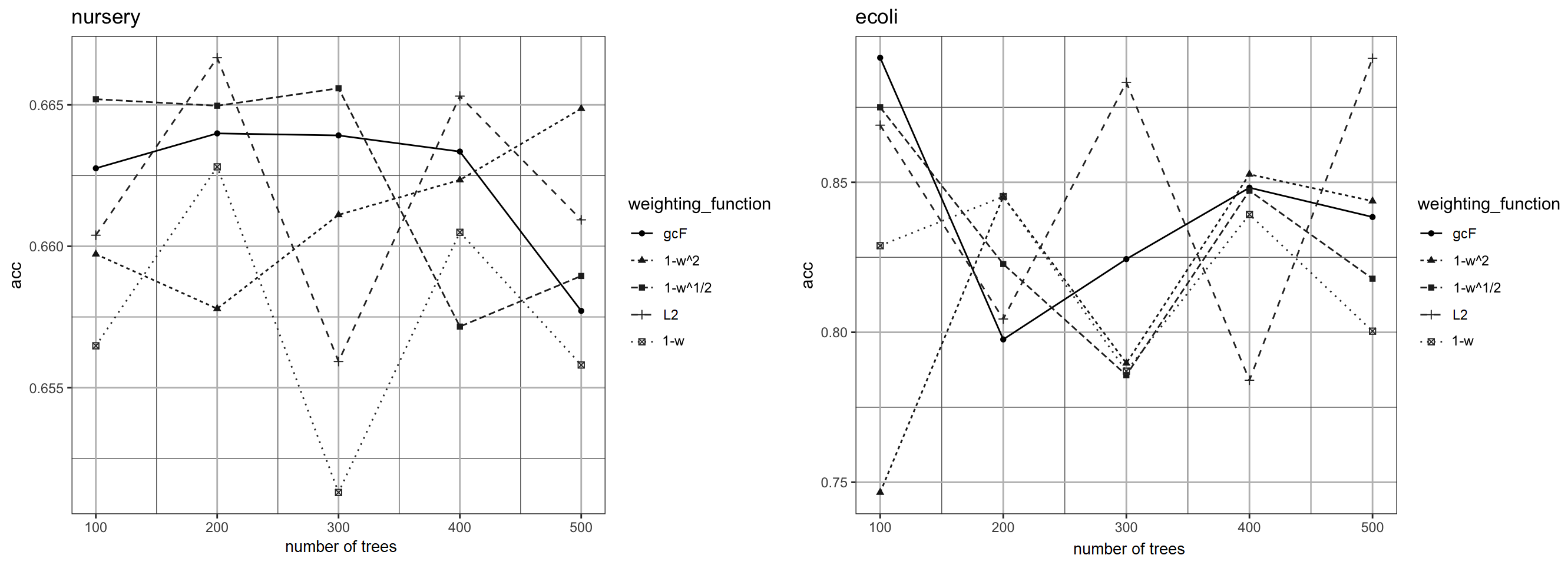}%
\caption{Accuracy measures as a function of the number of trees for the
Nursery and Ecoli datasets}%
\end{center}
\end{figure}
%EndExpansion
%

%TCIMACRO{\FRAME{ftbpFU}{5.2676in}{1.9207in}{0pt}{\Qcb{Accuracy measures as a
%function of the number of trees for the Dermatology and IMDB datasets}%
%}{\Qlb{fig:dermat_IMDB}}{dermatology-imdb.png}%
%{\special{ language "Scientific Word";  type "GRAPHIC";
%maintain-aspect-ratio TRUE;  display "USEDEF";  valid_file "F";
%width 5.2676in;  height 1.9207in;  depth 0pt;  original-width 13.7496in;
%original-height 5.0004in;  cropleft "0";  croptop "1";  cropright "1";
%cropbottom "0";  filename 'Dermatology-IMDB.png';file-properties "XNPEU";}} }%
%BeginExpansion
%EndExpansion

Tabl. \ref{t:AWDF2} shows accuracy measures in accordance with the first
strategy of using weights when we randomly draw instances from the training
set with replacement in accordance with the probability distribution
corresponding to the weights. One can see from Tabl. \ref{t:AWDF2} that AWDF
is comparable with gcForest, but it yields the best accuracy on 7 out of 15
datasets tested. This can be explained by the fact that we significantly
reduce the training time by removing a large part of instances from the
training process after the first cascade level. Perhaps, a more fine tuning of
AWDF (choice of an appropriate function of weights) may improve the
classification results.%

%TCIMACRO{\TeXButton{B}{\begin{table}[tbp] \centering}}%
%BeginExpansion
\begin{table}[tbp] \centering
%EndExpansion
\caption{Accuracy measures gcForest and four cases of the weight function for the first strategy of using weights}%
\begin{tabular}
[c]{cccccc}\hline
Data set & gcF & $1-w^{2}$ & $1-w^{1/2}$ & $L_{2}$ & $1-w$\\\hline
Adult & $\mathbf{86.12}$ & $85.98$ & $85.72$ & $85.95$ & $85.84$\\\hline
Car & $98.28$ & $98.16$ & $98.27$ & $97.88$ & $\mathbf{98.39}$\\\hline
Diabet & $69.05$ & $68.57$ & $\mathbf{69.62}$ & $68.25$ & $68.38$\\\hline
EEG & $\mathbf{95.75}$ & $94.92$ & $95.29$ & $94.70$ & $95.13$\\\hline
Haberman & $\mathbf{74.15}$ & $72.17$ & $73.22$ & $73.35$ & $72.97$\\\hline
Ion & $\mathbf{94.32}$ & $93.94$ & $94.10$ & $94.21$ & $93.78$\\\hline
Seeds & $\mathbf{93.17}$ & $92.90$ & $92.09$ & $92.45$ & $91.88$\\\hline
Seismic & $93.50$ & $\mathbf{93.79}$ & $\mathbf{93.79}$ & $93.60$ &
$93.47$\\\hline
TAE & $52.76$ & $51.88$ & $52.38$ & $53.63$ & $\mathbf{53.65}$\\\hline
TTTE & $\mathbf{99.03}$ & $98.33$ & $98.81$ & $98.41$ & $98.63$\\\hline
Website & $\mathbf{90.15}$ & $90.03$ & $88.92$ & $89.86$ & $89.67$\\\hline
WCR & $72.81$ & $71.52$ & $72.12$ & $72.68$ & $\mathbf{73.33}$\\\hline
Letter & $\mathbf{97.14}$ & $96.47$ & $96.59$ & $96.40$ & $96.55$\\\hline
Nursery & $66.40$ & $\mathbf{66.55}$ & $66.31$ & $66.38$ & $66.44$\\\hline
Dermatology & $58.10$ & $\mathbf{65.22}$ & $60.68$ & $65.21$ & $58.15$\\\hline
\end{tabular}
\label{t:AWDF2}%
%TCIMACRO{\TeXButton{E}{\end{table}}}%
%BeginExpansion
\end{table}%
%EndExpansion

We again apply the $t$-test for the difference between accuracy measures of
gcForest and AWDF. The results of computing the $t$ statistics for the
difference between the best value of AWDF and gcForest are the p-values
denoted as $p$ and the $95\%$ confidence interval for the mean $0.465$, which
are $p=0.3577$ and $[-0.584,1.515]$, respectively. The $t$-test demonstrates
that there is not significant difference between accuracy measures of gcForest
and AWDF when we apply the first strategy of using weights.

\begin{figure}
[ptb]
\begin{center}
\includegraphics[
%natheight=5.000400in,
%natwidth=13.792000in,
height=2.0012in,
width=5.5002in
]%
{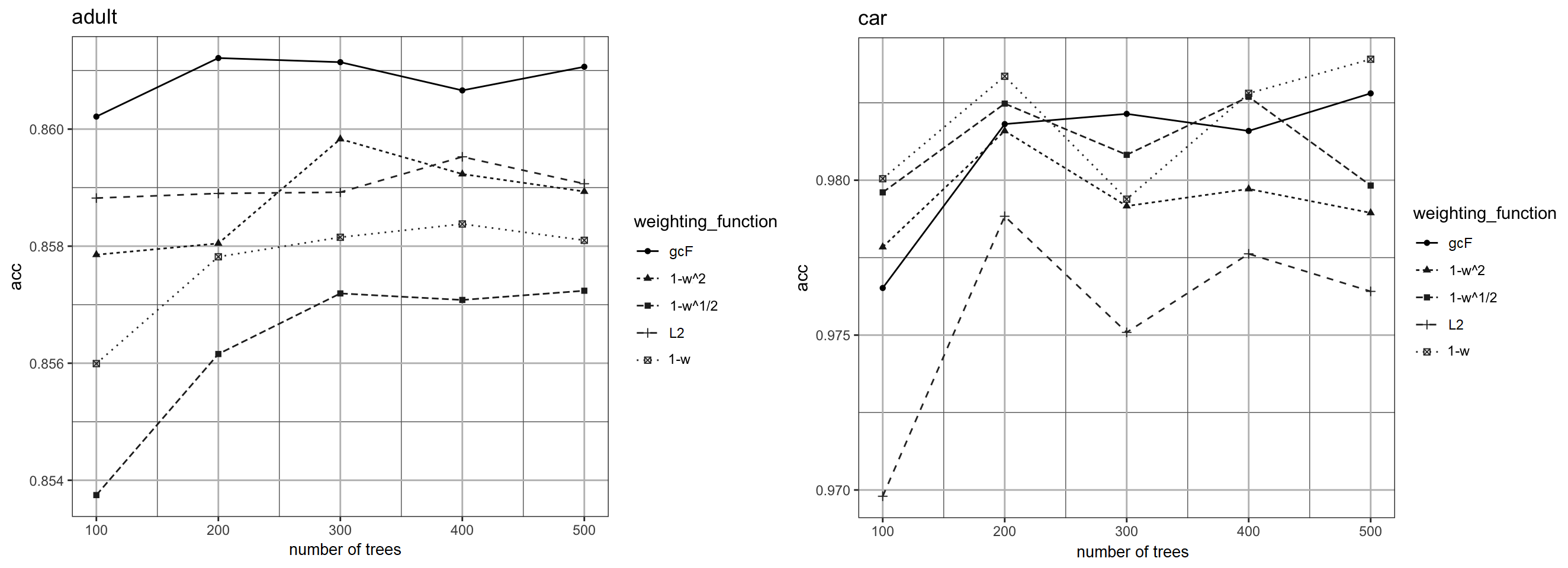}%
\caption{Accuracy measures as a function of the number of trees for the Adult
and Car datasets for the first strategy of using weights}%
\label{fig:as_car-2}%
\end{center}
\end{figure}

Examples of the dependencies of the accuracy measures on the number of
decision trees in every RF for the first strategy of using weights are shown
in Fig. \ref{fig:as_car-2}. We clearly see from Fig. \ref{fig:as_car-2} that
AWDF does not outperform gcForest.%

%TCIMACRO{\FRAME{ftbpFU}{5.5002in}{2.0012in}{0pt}{\Qcb{Accuracy measures as a
%function of the number of trees for the Adult and Car datasets for the first
%strategy of using weights}}{\Qlb{fig:as_car-2}}{adult-car-2.png}%
%{\special{ language "Scientific Word";  type "GRAPHIC";
%maintain-aspect-ratio TRUE;  display "USEDEF";  valid_file "F";
%width 5.5002in;  height 2.0012in;  depth 0pt;  original-width 13.792in;
%original-height 5.0004in;  cropleft "0";  croptop "1";  cropright "1";
%cropbottom "0";  filename 'Adult-Car-2.png';file-properties "XNPEU";}} }%
%BeginExpansion
%EndExpansion

It follows from the above results that the second strategy of using weights
provides better accuracies in comparison with the first strategy.

\begin{figure}
[ptb]
\begin{center}
\includegraphics[
%natheight=5.000400in,
%natwidth=13.765200in,
height=2.0332in,
width=5.578in
]%
{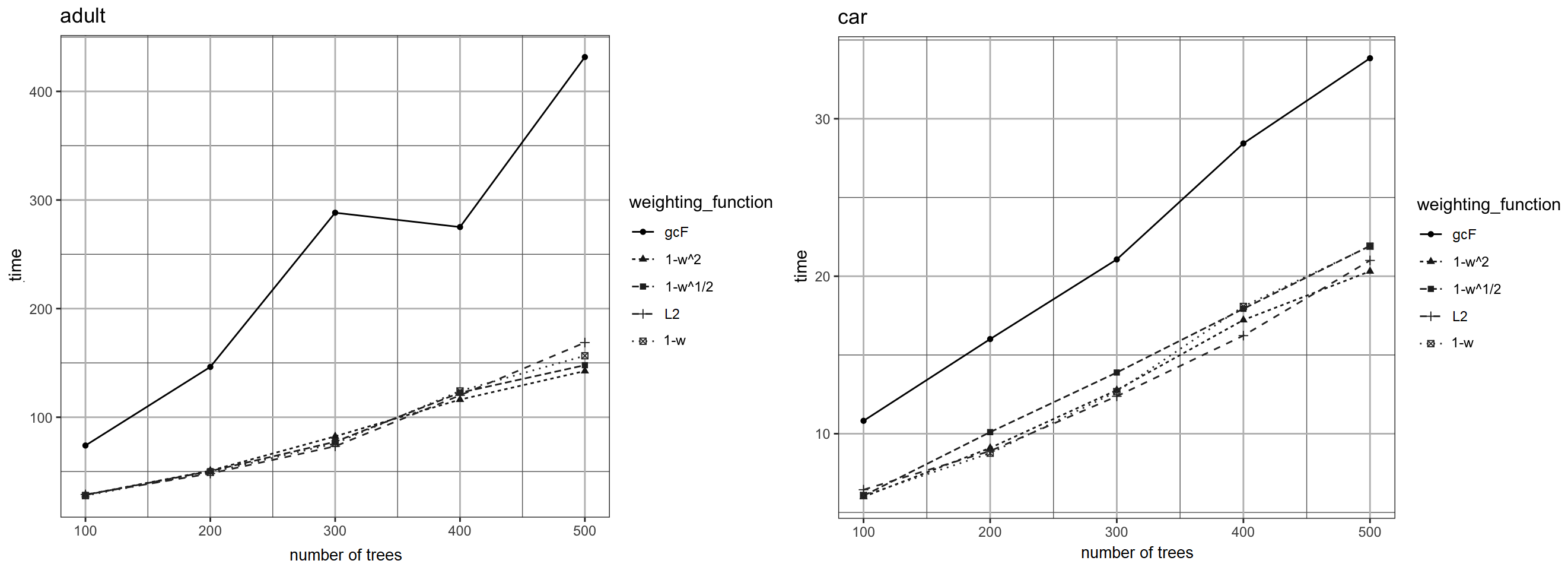}%
\caption{The training time of classifiers as a function of the number of trees
for the Adult and Car datasets for the second strategy of using weights and
threshold $\eta_{q}=0.95$}%
\label{fig:as_car-3}%
\end{center}
\end{figure}

Let us consider how the introduction of threshold $\eta_{q}$ for AWDF impacts
on the accuracy measures. Tabl. \ref{t:AWDF3} shows accuracy measures in
accordance with the second strategy of using weights by $\eta_{q}=0.95$. One
can see from Tabl. \ref{t:AWDF3} that AWDF yields the best accuracy on 13 out
of 16 datasets tested. At the same time, we have to note that the training
time by using the threshold is significantly reduced. Fig. \ref{fig:as_car-3}
shows examples of the training time as a function of the number of trees. One
can see that the training time is reduced for AWDF in comparison with
gcForest. %

%TCIMACRO{\TeXButton{B}{\begin{table}[tbp] \centering}}%
%BeginExpansion
\begin{table}[tbp] \centering
%EndExpansion
\caption{Accuracy measures gcForest and four cases of the weight function for the second strategy of using weights and threshold $\eta _{q}=0.95$}%
\begin{tabular}
[c]{cccccc}\hline
Data set & gcF & $1-w^{2}$ & $1-w^{1/2}$ & $L_{2}$ & $1-w$\\\hline
Adult & $86.12$ & $86.09$ & $86.10$ & $86.04$ & $\mathbf{86.13}$\\\hline
Car & $\mathbf{98.28}$ & $96.46$ & $96.83$ & $96.16$ & $96.54$\\\hline
Diabet & $69.05$ & $69.71$ & $69.71$ & $\mathbf{69.92}$ & $69.84$\\\hline
EEG & $\mathbf{95.75}$ & $92.54$ & $92.44$ & $92.45$ & $92.47$\\\hline
Haberman & $74.15$ & $74.52$ & $73.47$ & $73.10$ & $\mathbf{73.59}$\\\hline
Ion & $94.32$ & $\mathbf{95.18}$ & $94.26$ & $94.91$ & $93.51$\\\hline
Seeds & $93.17$ & $93.26$ & $92.90$ & $\mathbf{93.71}$ & $93.17$\\\hline
Seismic & $93.50$ & $93.37$ & $93.68$ & $\mathbf{93.73}$ & $93.54$\\\hline
TAE & $52.76$ & $53.26$ & $\mathbf{54.39}$ & $53.88$ & $53.38$\\\hline
TTTE & $99.03$ & $97.42$ & $\mathbf{97.62}$ & $97.20$ & $97.22$\\\hline
Website & $90.15$ & $90.27$ & $90.32$ & $90.25$ & $\mathbf{90.52}$\\\hline
WCR & $72.81$ & $72.60$ & $\mathbf{72.94}$ & $72.16$ & $72.21$\\\hline
Letter & $\mathbf{97.14}$ & $96.35$ & $96.35$ & $96.35$ & $96.35$\\\hline
Yeast & $63.03$ & $62.97$ & $63.00$ & $\mathbf{63.11}$ & $63.04$\\\hline
Nursery & $66.40$ & $65.24$ & $65.51$ & $66.13$ & $\mathbf{65.95}$\\\hline
Dermatology & $58.10$ & $\mathbf{58.98}$ & $58.39$ & $58.64$ & $57.99$\\\hline
\end{tabular}
\label{t:AWDF3}%
%TCIMACRO{\TeXButton{E}{\end{table}}}%
%BeginExpansion
\end{table}%
%EndExpansion
%

%TCIMACRO{\FRAME{ftbpFU}{5.578in}{2.0332in}{0pt}{\Qcb{The training time of
%classifiers as a function of the number of trees for the Adult and Car
%datasets for the second strategy of using weights and threshold $\eta
%_{q}=0.95$}}{\Qlb{fig:as_car-3}}{adult-car-3.png}%
%{\special{ language "Scientific Word";  type "GRAPHIC";
%maintain-aspect-ratio TRUE;  display "USEDEF";  valid_file "F";
%width 5.578in;  height 2.0332in;  depth 0pt;  original-width 13.7652in;
%original-height 5.0004in;  cropleft "0";  croptop "1";  cropright "1";
%cropbottom "0";  filename 'Adult-Car-3.png';file-properties "XNPEU";}}}%
%BeginExpansion
%EndExpansion
\qquad

\section{Conclusion}

The proposed modification of the confidence screening mechanism based on
adaptive weighing of every training instance at each cascade level of DF has
demonstrated good performance in comparison with gcForest by means of
numerical experiments on several datasets. Its implementation is very simple
and is similar to the well-known AdaBoost model in a sense that it updates
weights of training instances at each level of the forest cascade.

The idea underlying AWDF is very simple and its implementation does not
require a large additional time because weights are computed in a simple way
without solving optimization problems.

Similarly to gcForest, the main advantage of AWDF is that it opens a door for
developing many new adaptive weight models which could take into account
different rules for updating and assigning the weights to instances at
different levels of the cascade. Moreover, the weights can be assigned in
accordance with the problem solved, for example, for improving the DF transfer
learning algorithms, for improving the distance metric learning algorithms,
etc. In other words, the weights can control the DF models. The development of
the corresponding algorithms is a problem for further research.

It should be noted that the proposed approach for adaptive weighing of every
training instances at each cascade level can be simply extended on cases when
classifiers different from RFs are used at every level because there are
weighted versions of the most classifiers. The choice of optimal structures is
also a problem for further research.

\section*{Acknowledgement}

This work is supported by the Russian Science Foundation under grant 18-11-00078.

%%\bibliographystyle{plain}
%%\bibliography{Autoencoder,Boosting,Classif_bib,Deep_Forest,Interv_NN,IntervalClass,MYBIB,MYUSE,Robots,Transf_Learn}

\end{document}